\documentclass[sigconf]{acmart}

\usepackage{amsmath}
\usepackage{cleveref} 
\AtBeginDocument{
  }

\usepackage{booktabs, adjustbox, multirow, array}
\usepackage{algorithm}
\usepackage{algorithmic}

\renewcommand\footnotetextcopyrightpermission[1]{} 
\begin{document}

\title{SAGE: Sequence-level Adaptive Gradient Evolution for Generative Recommendation}

\author{Yu Xie}
\email{zanghai1@xiaohongshu.com}

\author{Xingkai Ren}
\email{lingtong1@xiaohongshu.com}

\author{Ying Qi}
\email{yingqi1@xiaohongshu.com}

\author{Yao Hu}
\email{huyao@xiaohongshu.com}

\begin{abstract}
  Reinforcement learning–based preference optimization is increasingly used to align list-wise generative recommenders with complex, multi-objective user feedback, yet existing optimizers such as Gradient-Bounded Policy Optimization (GBPO) exhibit structural limitations in recommendation settings. We identify a Symmetric Conservatism failure mode in which symmetric update bounds suppress learning from rare positive signals (e.g., cold-start items), static negative-sample constraints fail to prevent diversity collapse under rejection-dominated feedback, and group-normalized multi-objective rewards lead to low-resolution training signals. To address these issues, we propose SAGE (Sequence-level Adaptive Gradient Evolution), a unified optimizer designed for list-wise generative recommendation. SAGE introduces sequence-level signal alignment via a geometric-mean importance ratio and a decoupled multi-objective advantage estimator to reduce token-level variance and mitigate reward collapse, together with asymmetric adaptive bounding that applies positive Boost updates to successful slates and an entropy-aware penalty to discourage low-diversity failures. Experiments on Amazon Product Reviews and the large-scale RecIF-Bench demonstrate consistent improvements in top-K accuracy, cold-start recall, and diversity across both Semantic-ID and native-text action spaces, while preserving numerical stability during training. These results suggest that asymmetric, sequence-aware policy optimization provides a principled and effective framework for addressing optimization failures in generative recommendation.
\end{abstract}


\maketitle

\section{Introduction}

In the digital age, recommender systems have become indispensable tools for user navigation of vast information landscapes. However, traditional methodologies, primarily leveraging historical interaction data and techniques such as collaborative filtering and matrix factorization, exhibit inherent shortcomings. Although effective in data-rich environments for capturing certain facets of user preference, these approaches often overlook valuable external knowledge, thereby contributing to the well-established "filter bubble" effect \cite{koren2009matrix}. Moreover, industrial deployments frequently maintain a separation between model optimization and strategic iteration. This division, often manifested through distinct teams dedicated to model tuning versus strategy adjustment \cite{covington2016deep, liu2017cascade, qin2022rankflow, luo2024integrating}, can impede the rapid adaptation necessary to address evolving business requirements.

Recent advancements in large-scale pre-trained language models (LLMs) have introduced a transformative opportunity for the recommender system domain. These models demonstrate remarkable capabilities in semantic understanding and knowledge reasoning, and are capable of learning new tasks with limited data~\cite{brown2020language, devlin2019bert, luo2024integrating, li2023gpt4rec, yang2023palr, liao2024llara, ji2024genrec, li2025survey, deng2025onerec, wei2022chain, kojima2022large}. Capitalizing on this potential to reshape data construction and model architecture in recommender systems holds the promise of overcoming the inherent limitations of traditional approaches, leading to enhancements in both the accuracy and diversity of personalized recommendations.

In generative recommendation systems such as OneRec, adapting an LLM to the recommendation task typically requires constructing a dedicated vocabulary to convert videos into Semantic IDs. Although this design alleviates the issues of unbounded vocabulary growth and cold-start problems inherent in traditional ID-based recommendation, it still faces several practical challenges ~\cite{zhou2025onerec, zhou2025openonerec, rajput2023recommender}. 

First, \textbf{semantic collision} is a major issue: multiple distinct videos may be mapped to the exact same token sequence, while a large portion of tokens remain unmapped. This can lead to collision rates exceeding 30\%, meaning that the model is unable to distinguish 30\% of the items, resulting in severe information loss ~\cite{zhou2025openonerec}. Second, in industrial-scale scenarios where hundreds of millions of new videos are uploaded daily, maintaining the tokenizer incurs substantial operational overhead. 

Moreover, \textbf{LLM adaptation training is mandatory}. A dedicated training stage is required for the LLM to understand item tokens and absorb collaborative signals. General-purpose LLMs cannot interpret these newly defined ''Itemic Tokens'' (video IDs); thus, vocabulary expansion and alignment must be performed during the pre-training stage ~\cite{zhou2025onerec, zhou2025openonerec, artetxe2020cross}. This includes initializing embeddings for the new tokens and freezing other parameters during training so that the LLM can learn the semantics of video tokens.

To evaluate whether the optimization issues are intrinsic to the optimizer rather than a specific item tokenization, we consider two representative action spaces in generative recommendation: (i) Semantic-ID generation as in OneRec, and (ii) native-text item rendering without a separate item vocabulary. The approach is outlined as follows:

  \paragraph{System instantiation for evaluation: data construction.}
The core advantage of this paradigm lies in the fact that it can directly adopt the architecture of large language models without the need to construct a separate tokenization vocabulary, thus enabling the most efficient reuse of open-source large language models.
User-side: We integrate user profiles with historical behavior information, constructing prompt inputs that combine static user attributes and dynamic interaction data. This ensures that the large model can capture both the basic characteristics and behavioral habits of the user.
Item-side: We comprehensively leverage item descriptions, titles, and content-derived tags to provide a multi-faceted semantic representation of the items. Furthermore, we generate positive samples by constructing an item sequence based on the user’s subsequent interactions, guided by these tags.
Label Construction: We generate the training label, an item sequence that accurately reflects the user's preferences, based on their subsequent behavior data. This provides a more precise supervisory signal for model training.
  \paragraph{System instantiation for evaluation: two-stage training.}
SFT Stage: In the Supervised Fine-Tuning (SFT) stage, we employ a distilled version of Qwen3 to fine-tune the large language model using the constructed prompt data derived from recommender system user interactions. This process activates the model’s inherent capabilities for recommendation tasks, facilitating knowledge transfer and initial model adaptation. Additionally, high-quality samples generated in the RLHF stage will be added to the training samples.
RLHF Stage: Building upon the SFT stage, we propose SAGE (Sequence-level Adaptive Gradient Evolution), a unified optimization framework tailored for list-wise generative recommendation. SAGE introduces two key innovations: (1) Sequence-level Signal Decoupling: We propose a geometric mean importance ratio combined with a decoupled multi-objective advantage estimator. This eliminates token-level variance and resolves "Reward Collapse" in multi-objective scenarios (e.g., balancing clicks and duration). (2) Asymmetric Adaptive Dynamics: SAGE constructs a dynamic gradient manifold that is asymmetric by design. It introduces a "Boost Factor" to allow super-linear updates for high-potential cold-start items, and an "Entropy-Aware Penalty" that dynamically intensifies gradients for low-diversity negative samples to break information cocoons. Theoretical analysis and empirical results demonstrate that SAGE effectively unblocks cold-start traffic and sustains recommendation diversity, all while retaining the numerical stability of GBPO.

\begin{figure}
    \centering
    \includegraphics[width=0.9 \linewidth]{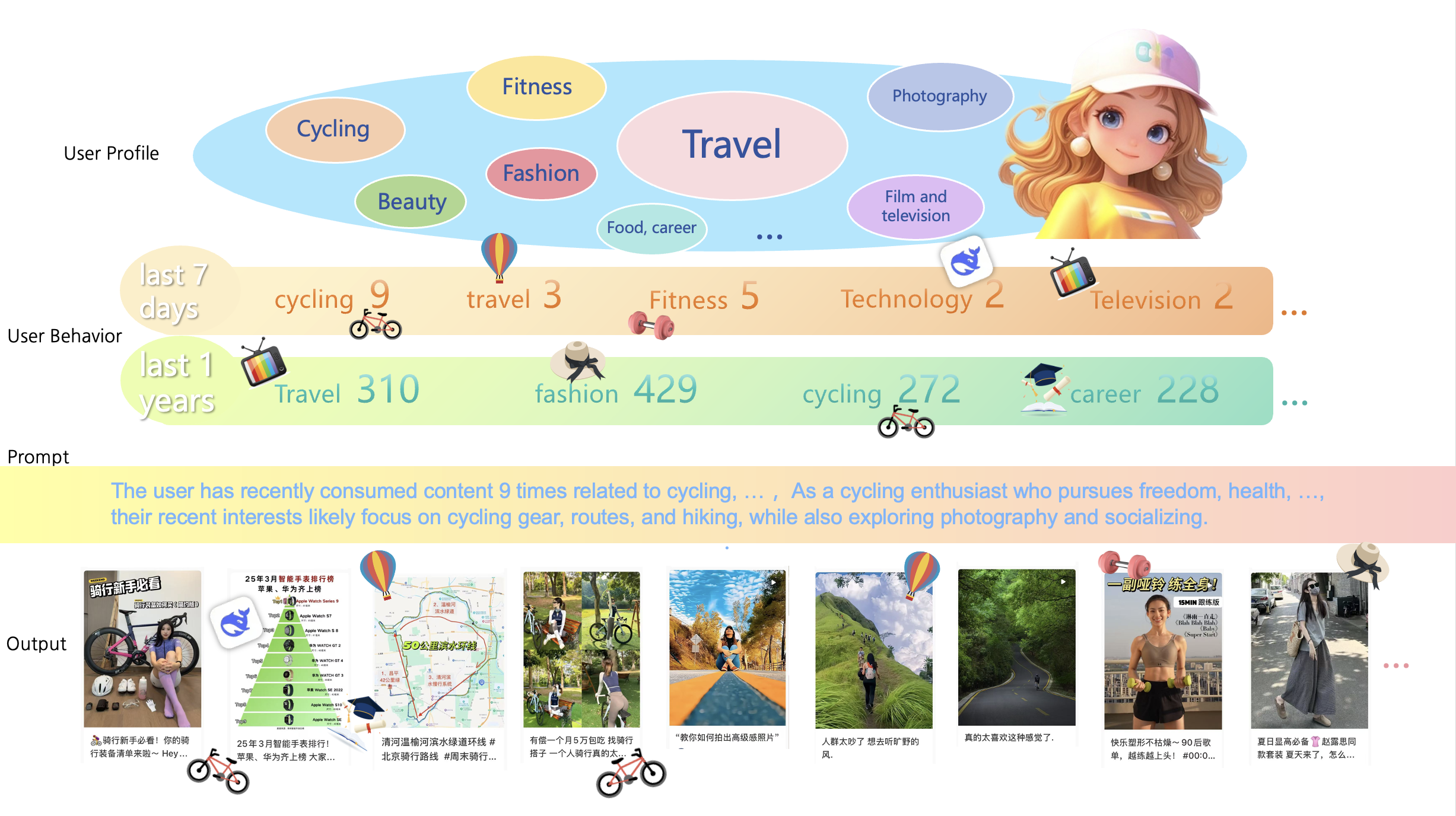}
    \caption{Illustrates the motivation behind our research}
    \Description{Conceptual illustration of the proposed approach and its motivation for combining large language models with reinforcement learning for recommendation.}
    \label{fig:result-pic}
\end{figure}

\paragraph{Contributions.}
We identify a practical failure mode of GBPO in generative recommendation, termed \emph{Symmetric Conservatism}, where symmetric conservative bounds suppress rare-item positive learning signals and can trigger diversity collapse under rejection-dominated feedback. To address this, we propose \textbf{SAGE}, a sequence-level, multi-objective policy optimization framework for list-wise generative recommendation. SAGE combines a sequence-level importance ratio for variance reduction, an asymmetric adaptive bounding rule with a positive boost and an entropy-aware penalty, and a decoupled normalization scheme for multi-objective rewards. We evaluate SAGE on both academic and industrial benchmarks and observe consistent gains in accuracy, cold-start discovery, and diversity, while maintaining stable training.

Overall, Figure~\ref{fig:result-pic} illustrates the motivation behind our research. Our approach offers a novel perspective on recommender system design by seamlessly integrating the strengths of large language models and reinforcement learning. It effectively addresses the limitations of traditional methods—particularly the issue of information silos—while providing a solid technical foundation for multi-objective and customized recommendation in real-world business scenarios.

\section{Related Work}
\label{sec: Related Work}

\subsection{Generative Recommendation and Its Challenges}
With the breakthroughs in generative artificial intelligence, the paradigm of recommender systems is shifting from traditional \textit{Discriminative Ranking} to \textit{Generative Recommendation}. The core idea of generative recommendation is to reformulate the task as an end-to-end autoregressive generation problem, directly generating the Semantic ID sequence of the target item, thereby enabling direct optimization toward the final objective. This requires a tokenization strategy that maps items into discrete token sequences (Semantic IDs). \textbf{OneRec} is a pioneering paradigm in this direction, relying on predefined structured vocabularies ~\cite{li2025survey, deng2025onerec}.

However, this paradigm introduces a dependency on a heavy vocabulary: the model must memorize the mapping between semantic tokens and specific items. The discrete nature of these tokens makes optimization difficult, as even slight deviations in token generation may lead to valid but irrelevant item IDs, or even entirely invalid sequences ~\cite{deng2025onerec, zhou2025onerec, ranzato2015sequence}.

\subsection{Evolution of Policy Optimization}
To align large models with complex business objectives, reinforcement learning–based methods have become mainstream. \textbf{GRPO} ~\cite{shao2024deepseekmath} eliminates the need for a Critic network by normalizing rewards within sampled groups, and it has been widely applied in reasoning tasks. However, GRPO suffers from the problem of \textit{Reward Collapse} in multi-objective optimization, where samples with different performance may end up with identical advantage values after normalization, reducing the resolution of training signals. In addition, negative feedback dominates recommendation scenarios, and directly applying standard RL often leads to gradient instability ~\cite{zhou2025onerec, zhao2017deep, afsar2022reinforcement, chen2019top}.

To address these issues, \textbf{OneRec-V2} ~\cite{zhou2025onerec} proposed \textbf{GBPO} (Gradient-Bounded Policy Optimization, OneRec-GBPO). GBPO introduces a dynamic gradient boundary derived from binary cross-entropy (BCE), effectively preventing gradient explosion caused by negative samples and improving stability in large-scale training.

Although GBPO resolves the stability issue, its symmetric and conservative design ($r \le 1$) introduces new bottlenecks. Cold-start video views dropped by \textbf{44.7\%}, because the strict upper bound on gradients prevents strong updates on sparse positive signals for new items. Cluster density increased by \textbf{11.7\%}, as the static boundary fails to preserve entropy-increasing exploratory behavior.

Therefore, we propose \textbf{SAGE}, a unified optimizer that is \textbf{sequence-level, adaptive, and multi-objective}.Token-level optimization introduces noise in Semantic ID generation. Following the logic of \textbf{GSPO} ~\cite{zheng2025group}, SAGE elevates the importance ratio to the sequence level (geometric mean), ensuring the overall consistency of tokenized item IDs. \textbf{BAPO} ~\cite{xi2025bapo} shows that standard clipping suppresses exploration. SAGE adopts BAPO's \textit{Entropy-Clip Rule} but transforms it into a dynamic penalty term for negative samples, countering the diversity collapse observed in OneRec-GBPO. To offset GBPO's suppression of new items, SAGE incorporates the \textit{Clip-Higher} philosophy from \textbf{DAPO} ~\cite{yu2025dapo}, enabling super-linear updates for positive cold-start signals. To address the reward collapse issue in GRPO, SAGE integrates the decoupled normalization strategy from \textbf{GDPO} ~\cite{liu2026gdpo}, achieving precise alignment across multiple objectives (e.g., clicks and Comment).

\paragraph{Positioning of SAGE.}
Conceptually, SAGE belongs to the PPO/GBPO family of bounded policy optimization methods, but differs in two aspects that are crucial for recommendation: (i) SAGE performs \emph{sequence-level} importance weighting to match list-wise feedback and reduce token-level variance; (ii) SAGE introduces an \emph{asymmetric, sample-adaptive} bounding rule---boosting positive successful slates while penalizing low-entropy failures more strongly---to address cold-start suppression and diversity collapse. For multi-objective feedback, SAGE adopts a decouple-then-aggregate advantage estimator to avoid reward collapse after group normalization.

\section{Method}
\label{sec:Method}

\subsection{Problem Setup}
We study list-wise generative recommendation as a slate generation problem ~\cite{ie2019slateq}.
Given a user context $c$ (e.g., profile and recent behaviors), a policy $\pi_{\theta}$ generates a slate $S=(i_1,\dots,i_L)$ autoregressively.
User feedback is provided at the slate level and can be multi-objective, denoted by $\mathbf{R}(S)=\{R_m(S)\}_{m=1}^{M}$.
Our goal is to optimize the expected utility under mixed feedback while maintaining stable training under rejection-dominated data:
\begin{equation}
\max_{\theta}\;\mathbb{E}_{S\sim\pi_{\theta}(\cdot|c)}\left[\sum_{m=1}^{M} w_m R_m(S)\right].
\end{equation}

\begin{figure}
    \centering
    \includegraphics[width=0.8\linewidth]{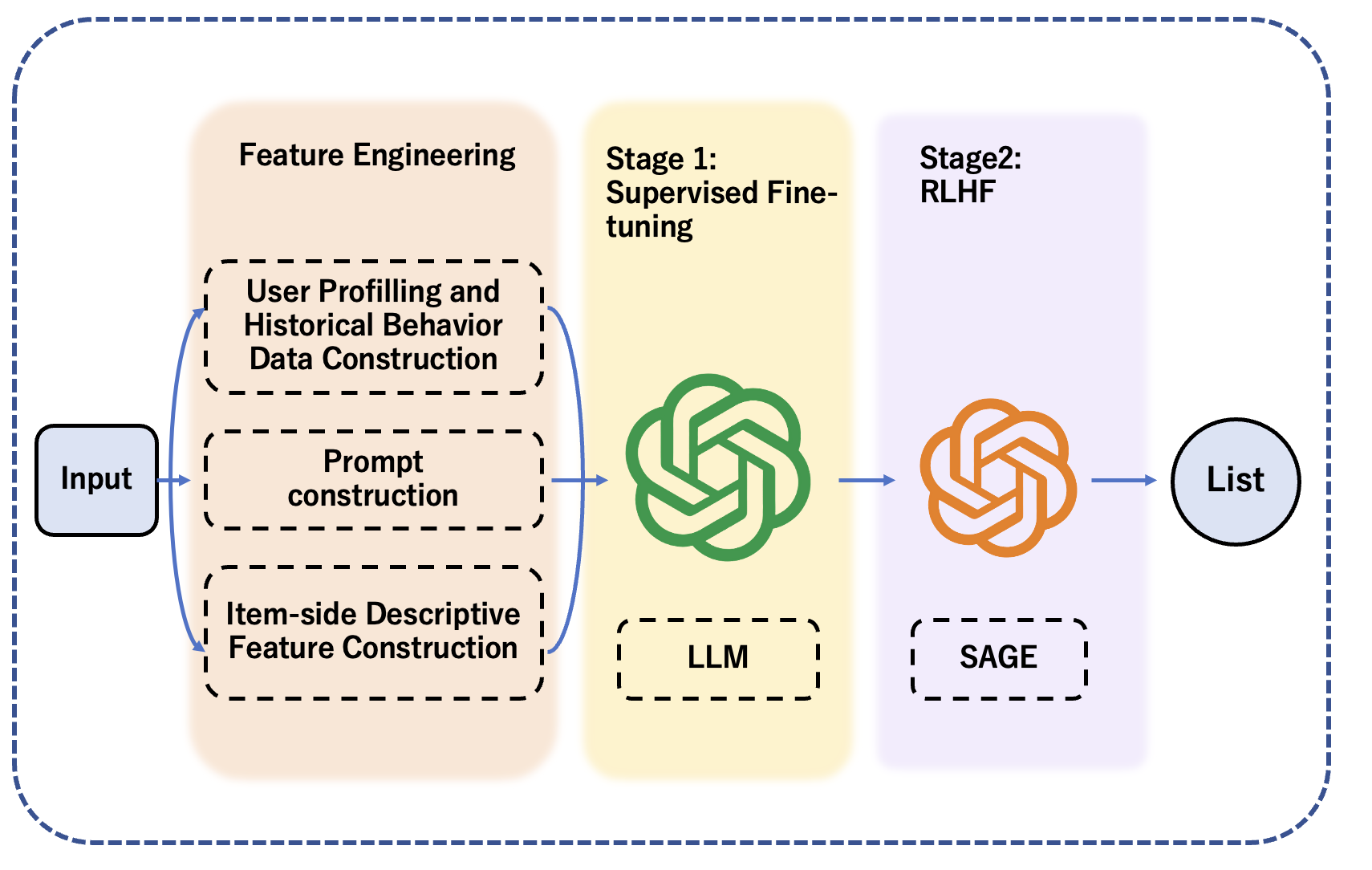}
    \caption{An overview of the modeling approach used in TextRec.}
    \label{fig: modeling approach}
\end{figure}

We instantiate the proposed formulation using a two-stage training pipeline (SFT followed by RLHF) on textualized recommendation data. The central algorithmic contribution is the RLHF optimizer, SAGE, which we describe in detail next and illustrate in Figure ~\ref{fig: modeling approach}.

\subsection{Implementation: Data Construction}

To align with the input requirements of large language models, we constructed a training dataset specifically tailored for recommender systems by transforming features into natural language descriptions.
\textbf{User-side Data}: We transformed user profiles (e.g., age, gender, interests) and recent interaction history (e.g., sequences of clicked, viewed, liked, bookmarked, or purchased items) into chronologically ordered textual descriptions, aggregated by topic. For instance, "Over the past 7 days, the user viewed "Sichuan Cuisine" videos twice and bookmarked one article titled "Chengdu Travel Guide".
\textbf{Item-side Data}: We primarily utilized item descriptions, titles, and granular tag labels derived through content understanding, alongside statistical item features, to generate natural language descriptions.
\textbf{Label Construction}: Based on user engagement metrics such as browsing duration and interaction behaviors (e.g., bookmarking, sharing, commenting, following, liking), we constructed a sequence of items that reflects the user's eventual preferences as positive training samples. This ensures the authenticity and business relevance of the labels.
\textbf{Prompt Construction}: We concatenated the user-side and item-side textual information. The objective is to explicitly recommend the most relevant sequence of items to the user, with candidate items drawn from the provided data pool.

Through this methodology, we transformed traditional structured and semi-structured recommendation data into a natural language format interpretable and processable by large language models (LLMs). This data construction approach not only preserves the benefits of traditional interaction data but also imbues large language models with rich semantic context, thereby mitigating the limitations of the "filter bubble" effect.

\subsection{Implementation: SFT Warm-up}

Following the construction of the textualized recommendation dataset, we proceeded to the Supervised Fine-Tuning (SFT) stage. The primary objective of this phase was not to immediately achieve optimal recommendation performance, but rather to "activate" or "cold-start" the pre-trained large language model (LLM), endowing it with the foundational ability to address recommendation tasks.

For efficiency, we used the pre-trained Qwen3-8B model and adopted the previously constructed dataset that is compact in scale yet high in quality, which includes high-quality sample data generated in the RLHF stage. In this stage, the emphasis was on leveraging a minimal amount of data to rapidly unlock the LLM’s potential within the recommendation domain, enabling it to learn the requisite input-output formats and develop a rudimentary understanding of recommendation task modeling. This process prepares the model for the subsequent, more intricate multi-step optimization.

Essentially, the SFT stage serves as a "warm-up" providing the model with baseline recommendation capabilities prior to reinforcement learning. While SFT can already address certain recommendation needs—particularly cold-start scenarios, owing to the LLM’s general knowledge—it is inherently insufficient for optimizing complex long-term objectives, balancing multiple recommendation facets (such as diversity and novelty), or executing advanced reasoning. Therefore, SFT acts as a critical precursor, priming the model for comprehensive optimization in the subsequent RLHF stage. This design closely mirrors the training paradigm of DeepSeek R1.

\subsection{RLHF Stage: Preference Alignment with Real-World User Interactions}
After establishing foundational capabilities in the Supervised Fine-Tuning (SFT) stage, we introduce the RLHF phase of model. To address the cold-start and diversity issues mentioned in OneRec-V2 (OneRec-GRPO), we propose SAGE (Sequence-level Adaptive Gradient Evolution).
The core contribution of SAGE is the \textbf{Asymmetric Dynamic Bounding} mechanism, which adjusts the optimization trajectory based on the nature of the feedback (Positive vs. Negative) and the entropy of the generated list.

\subsubsection{Background: GBPO in OneRec-V2}
\label{sec:gbpo_background}

We briefly review Gradient-Bounded Policy Optimization (GBPO) proposed in the OneRec-V2 Technical Report~\cite{zhou2025onerec}.
GBPO is a bounded policy optimization method tailored for recommendation settings with rejection-dominated feedback. It controls update magnitude by restricting the effective importance ratio to a conservative interval, which improves numerical stability compared to vanilla policy gradient.
However, the symmetric and conservative design of GBPO can induce a failure mode that we term \emph{Symmetric Conservatism}: (i) the upper bound suppresses learning from rare positive slates (e.g., cold-start items), and (ii) static constraints on negative updates may fail to prevent diversity collapse under repeated rejections.
These limitations motivate our asymmetric and entropy-aware extension, SAGE.

\subsubsection{Adaptive Gradient Evolution}

\paragraph{SAGE as asymmetric, sample-adaptive clipping.}
Let $r_{\text{slate}}(\theta)$ be the sequence-level importance ratio (defined later).
SAGE optimizes a PPO-style surrogate with an \emph{asymmetric} and \emph{entropy-aware} clipping rule:
\begin{equation}
\mathcal{L}_{\text{SAGE}}(\theta)=
\mathbb{E}\left[\mathrm{clip}_{\text{SAGE}}\!\left(r_{\text{slate}}(\theta), A_{\text{SAGE}}(S), H(S)\right)\cdot A_{\text{SAGE}}(S)\right],
\end{equation}
where $\mathrm{clip}_{\text{SAGE}}(\cdot)$ boosts positive updates up to $1+\epsilon_{\text{boost}}$ and applies stronger penalties to low-entropy rejected slates.

\paragraph{SAGE Objective Function.}
To address the scale inconsistency between the geometric mean ratio (token-scale) and the sum of log-probabilities (sequence-scale), we introduce a length-normalization term $1/L$:
\begin{equation}
\begin{split}
\nabla \mathcal{J}_{SAGE}(\theta) &= \mathbb{E}_{\tau \sim \pi_{old}} \Bigg[ \frac{r_{slate}(\theta)}{\Phi(S, A)} \cdot A_{SAGE} \\
&\cdot \underbrace{\frac{1}{L} \sum_{t=1}^{L} \nabla \log \pi_\theta(i_t | u, S_{<t})}_{\text{Item-Averaged Gradient}} \Bigg]
\end{split}
\end{equation}
Here, the factor $1/L$ normalizes the gradient magnitude to match the scale of $r_{\mathrm{slate}}(\theta)$ (typically around $1$), avoiding an implicit preference toward longer sequences. The term $\Phi(S,A)$ is an adaptive dynamic denominator defined below.

\paragraph{Case A: Positive boost for cold-start (mitigating suppression).}
When the slate receives positive feedback ($A_{\mathrm{SAGE}}\ge 0$), standard GBPO effectively caps the update ratio at $r\le 1$, which can suppress cold-start items whose initial probability under $\pi_{\mathrm{old}}$ is extremely small. We introduce a boost factor
\begin{equation}
\Phi_{\mathrm{pos}} =
\begin{cases}
\frac{1}{1+\epsilon_{\mathrm{boost}}}, & \text{if } r_{\mathrm{slate}}(\theta) > 1+\epsilon_{\mathrm{boost}},\\
1, & \text{otherwise}.
\end{cases}
\end{equation}
With $\epsilon_{\mathrm{boost}}\in[0.2,0.5]$, once $r_{\mathrm{slate}}(\theta)$ exceeds $1+\epsilon_{\mathrm{boost}}$, the denominator is locked so that the effective gradient coefficient is maintained at $1+\epsilon_{\mathrm{boost}}$. This enables successful cold-start items to gain probability in a super-linear manner, providing sufficient momentum to pass the initial exposure threshold.

\paragraph{Case B: Diversity-aware penalty (mitigating collapse).}
When the slate receives negative feedback ($A_{\mathrm{SAGE}}<0$), we distinguish failed exploration from homogenized repetition by using the list entropy $H(S)$ as a proxy for semantic diversity:
\begin{equation}
\Phi_{\mathrm{neg}} = \max\!\left(1,\; \frac{1-r_{\mathrm{slate}}(\theta)}{\lambda(H(S))}\right).
\end{equation}
The regulation factor $\lambda$ is negatively correlated with diversity:
\begin{equation}
\lambda(H(S)) = 1 + \beta \cdot \tanh\!\left(\max(0,\bar{H}-H(S))\right),
\end{equation}
where $H(S)$ denotes the slate diversity score (e.g., category entropy), $\bar{H}$ is the moving average of historical diversity, and $\beta$ is the diversity temperature. If $H(S)<\bar{H}$, then $\lambda>1$, which reduces the denominator and yields a larger gradient magnitude, thereby imposing a stronger penalty on ``cocooning'' behavior. Conversely, if $H(S)\ge \bar{H}$, then $\lambda\approx 1$ and the update reverts to the standard GBPO boundary, treating the rejection as a regular training signal without excessive punishment.

\begin{figure}[t]
    \centering
    \includegraphics[width=0.9\linewidth]{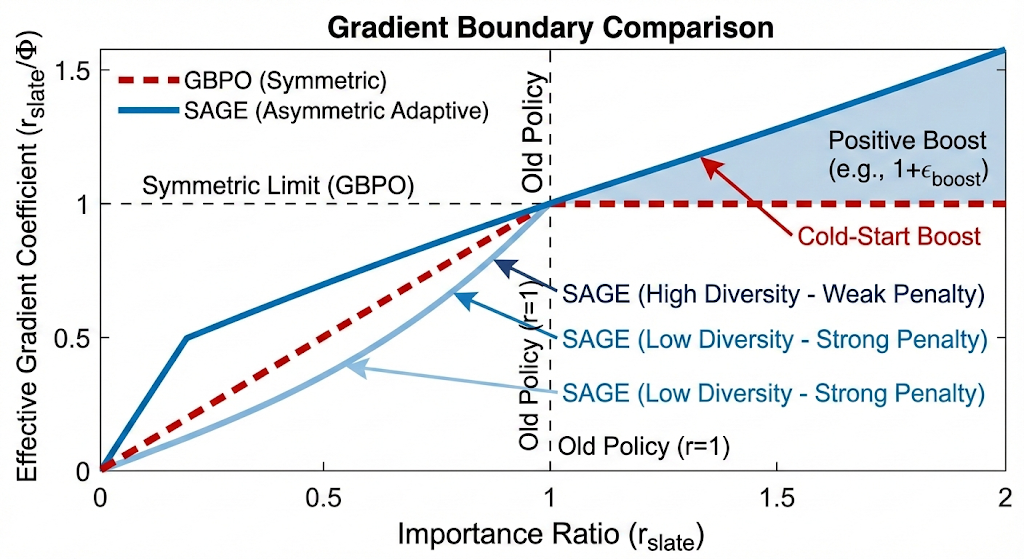} 
    \caption{\textbf{Comparison of Gradient Boundaries.} This figure illustrates the effective gradient coefficient ($r_{slate}/\Phi$) as a function of the importance ratio ($r_{slate}$). \textbf{GBPO (Dashed Red)} applies a static symmetric bound, capping positive updates at 1.0. In contrast, \textbf{SAGE (Solid Blue)} employs an asymmetric dynamic bound: it boosts high-potential positive samples beyond 1.0 (Case A) and applies entropy-aware penalties to negative samples (Case B), imposing stronger penalties on low-diversity slates (light blue) compared to high-diversity ones (dark blue).}
    \label{fig:gradient1}
\end{figure}

\begin{figure}[t]
    \centering
    \includegraphics[width=0.9\linewidth]{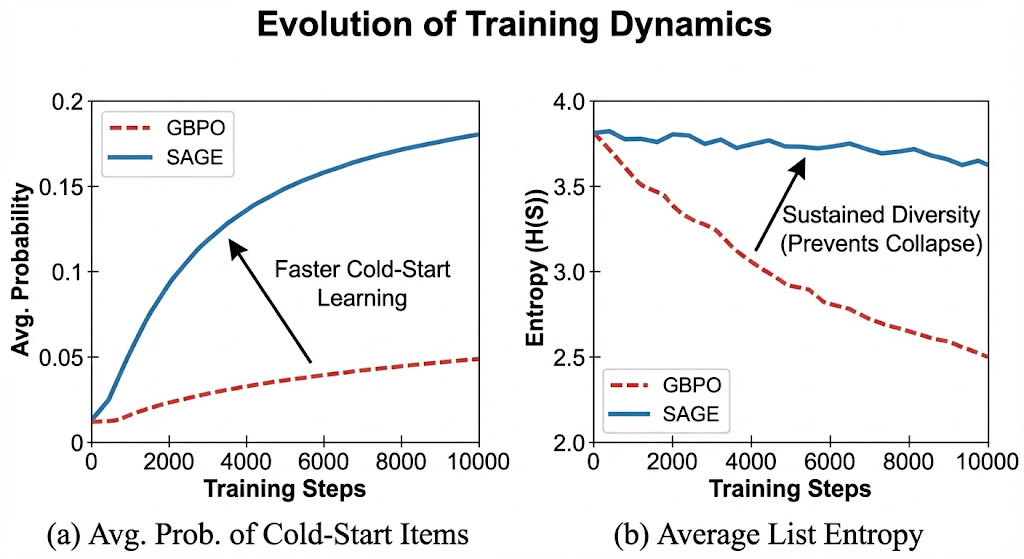} 
    \caption{\textbf{Evolution of Training Dynamics.} (a) SAGE rapidly boosts the generation probability of cold-start items compared to GBPO, demonstrating the effectiveness of the Positive Boost mechanism. (b) SAGE maintains consistently higher list diversity throughout training, verifying that the Entropy-Aware Penalty successfully prevents mode collapse, whereas GBPO tends to converge towards lower-entropy states.}
    \label{fig:training_dynamics}
\end{figure}

We illustrate the comparison of gradient boundaries in~\cref{fig:gradient1} and the evolution of training dynamics in~\cref{fig:training_dynamics}.

\paragraph{Algorithmic summary.}
The above design can be implemented as a PPO-style on-policy update with (i) a sequence-level importance ratio, (ii) decouple-then-aggregate multi-objective advantages, and (iii) an asymmetric, entropy-aware clipping rule.
For completeness, we summarize the training procedure in Algorithm~\ref{alg:sage_main}.

\begin{algorithm}[t]
\caption{SAGE: Sequence-level Adaptive Gradient Evolution}
\label{alg:sage_main_revised}
\begin{algorithmic}[1]
\REQUIRE Old policy $\pi_{\theta_{\text{old}}}$, trainable policy $\pi_{\theta}$, batch of contexts $\mathcal{B}$, slate length $L$, group size $K$, objectives $\{R_m\}_{m=1}^{M}$, boost $\epsilon_{\text{boost}}$, diversity temperature $\beta$, clip param $\epsilon_{\text{clip}}$
\ENSURE Updated parameters $\theta$

\STATE Initialize running diversity target $\bar{H}$

\FOR{each training step}
    \STATE \textbf{// Step 1: Sampling \& Evaluation}
    \STATE Sample batch $\mathcal{B} \sim \mathcal{D}$
    \STATE Generate $K$ slates $\{S_{u,k}\}$ for each user $u \in \mathcal{B}$ using $\pi_{\theta_{\text{old}}}$
    \STATE Compute Rewards $\{R_m(S_{u,k})\}$ and Diversity $H(S_{u,k})$
    
    \STATE \textbf{// Step 2: Signal Decoupling (Eq. 7)}
    \FOR{each user $u \in \mathcal{B}$}
        \FOR{each objective $m$}
            \STATE Normalize $R_m$ within group $K$: $\hat{R}_{u,k,m} \leftarrow \frac{R_{u,k,m} - \mu_{u,m}}{\sigma_{u,m} + \epsilon}$
        \ENDFOR
        \STATE Aggregate: $\hat{A}_{u,k} \leftarrow \sum_{m=1}^{M} w_m \cdot \hat{R}_{u,k,m}$
    \ENDFOR
    \STATE \textbf{Batch Normalize} $\hat{A}$ over entire batch $\mathcal{B}$ to get final $A_{u,k}$
    
    \STATE \textbf{// Step 3: Asymmetric Adaptive Optimization}
    \FOR{each sample $(u,k)$ in batch}
        \STATE Compute Sequence Ratio (Geo-Mean):
        \STATE $r_{u,k} \leftarrow \exp\left(\frac{1}{L}\sum_{t=1}^{L}\log \frac{\pi_{\theta}(i_{t}| c,S_{<t})}{\pi_{\theta_{\text{old}}}(i_{t}| c,S_{<t})}\right)$
        
        \IF{$A_{u,k} \ge 0$} 
            \STATE \textbf{Case A (Positive Boost):}
            \STATE Set lower bound: $\epsilon_{-} \leftarrow \epsilon_{\text{clip}}$
            \STATE Set upper bound: $\epsilon_{+} \leftarrow \epsilon_{\text{boost}}$ \COMMENT{Allow super-linear update}
        \ELSE
            \STATE \textbf{Case B (Diversity Penalty):}
            \STATE Calculate regulation: $\lambda \leftarrow 1 + \beta \cdot \tanh(\max(0,\bar{H}-H(S_{u,k})))$
            \STATE Set lower bound: $\epsilon_{-} \leftarrow \lambda - 1$ \COMMENT{Stronger penalty if low entropy}
            \STATE Set upper bound: $\epsilon_{+} \leftarrow \epsilon_{\text{clip}}$
        \ENDIF
        
        \STATE \textbf{Compute PPO Surrogate:}
        \STATE $r_{\text{clip}} \leftarrow \text{clip}(r_{u,k}, 1-\epsilon_{-}, 1+\epsilon_{+})$
        \STATE $\ell_{u,k} \leftarrow \min(r_{u,k} A_{u,k}, r_{\text{clip}} A_{u,k})$
    \ENDFOR

    \STATE $\mathcal{L}_{\text{SAGE}} \leftarrow -\frac{1}{|\mathcal{B}| \cdot K} \sum \ell_{u,k}$
    \STATE Update $\theta$ by descending $\nabla_\theta \mathcal{L}_{\text{SAGE}}$
    \STATE Update $\bar{H}$ via EMA
\ENDFOR
\end{algorithmic}
\end{algorithm}

\subsubsection{Multi-Objective Signal Decoupling}
\label{sec:mo_adv}
To construct a high-fidelity learning signal from complex user feedback (clicks, duration, comments), we adopt a \textit{Decouple-then-Aggregate} strategy inspired by GDPO.

\textbf{Decoupled Slate Advantage.} Directly summing weighted rewards can cause \textit{Reward Collapse}, where distinct behaviors map to identical advantage values. To avoid this, we normalize each objective independently within the sampling group and then aggregate them. For a slate $S$, the advantage is defined as
\begin{equation}
\begin{split}
A_{SAGE}(S) = \text{Norm}_{batch} \Bigg( \sum_{m=1}^{M} w_m \\
\cdot \frac{R_m(S) - \mu_m^{group}}{\sigma_m^{group} + \epsilon} \Bigg)
\end{split}
\end{equation}
where $R_m(S)$ denotes the raw reward of the $m$-th objective (e.g., Click or Comment), and $\mu_m^{group}, \sigma_m^{group}$ are the mean and standard deviation computed within the sampled group $G$ for the same user query. The operator $\text{Norm}_{batch}$ applies a final batch-level $z$-score normalization so that $A_{SAGE}$ stays on a stable scale (approximately standard normal), which helps maintain well-behaved gradient magnitudes.

\textbf{Sequence-level Ratio Formulation.} Since user feedback is directed at the item level, token-level importance ratios can introduce high variance. We therefore define the importance ratio at the slate level as the geometric mean of token probability ratios across the generated list:
\begin{equation}
\begin{split}
r_{slate}(\theta) = \exp \Bigg( \frac{1}{L} \sum_{t=1}^{L} \log \\
\frac{\pi_\theta(i_t | u, S_{<t})}{\pi_{\theta_{old}}(i_t | u, S_{<t})} \Bigg)
\end{split}
\end{equation}
with a fixed slate length $L=10$. \textit{Physical Meaning:} this formulation smooths intra-item token fluctuations and can be interpreted as the average probability gain of generating the \textit{entire slate} under the new policy relative to the old policy.

\begin{table*}[t]
\centering
\caption{Overall performance comparison on three Amazon Product Reviews datasets. Bold indicates the best performance, and underline indicates the second best.}
\label{tab:main_results1}
\resizebox{\textwidth}{!}{%

\begin{tabular}{c|cccc|cccc|cccc|cccc}
\hline
\multicolumn{1}{c|}{\textbf{Dataset}} & \multicolumn{4}{c|}{\textbf{Beauty}} & \multicolumn{4}{c|}{\textbf{Sports}} & \multicolumn{4}{c}{\textbf{Toys}} \\ \hline
\multicolumn{1}{c|}{\textbf{Method}} & R@5 & R@10 & N@5 & N@10 & R@5 & R@10 & N@5 & N@10 & R@5 & R@10 & N@5 & N@10  \\ \hline
SASRec & 0.0393 & 0.0639 & 0.0209 & 0.0289 & 0.0240 & 0.0389 & 0.0130 & 0.0178 & 0.0420 & 0.0658 & 0.0217 & 0.0294 \\
ReaRec & 0.0488 & 0.0702 & 0.0341 & 0.0409 & 0.0231 & 0.0348 & 0.0152 & 0.0189 & 0.0517 & 0.0706 & 0.0369 & 0.0430 \\
TIGER & 0.0413 & 0.0628 & 0.0277 & 0.0346 & 0.0216 & 0.0331 & 0.0145 & 0.0182 & 0.0367 & 0.0527 & 0.0255 & 0.0307 \\
LC-Rec & 0.0495 & 0.0764 & 0.0338 & 0.0424 & 0.0269 & 0.0418 & 0.0177 & 0.0225 & 0.0350 & 0.0549 & 0.0221 & 0.0285 \\ 
OneRec-GRPO & 0.0646 & 0.0924 & 0.0456 & 0.0545 & 0.0365 & 0.0547 & 0.0252 & 0.0310 & 0.0693 & 0.0953 & 0.0496  & 0.0579 \\
OneRec-GBPO & 0.0649 & 0.0928 & 0.0461 & 0.0550 & 0.0368 & 0.0551 & 0.0255 & 0.0317 & 0.0698 & 0.0958 & 0.0499  & 0.0581 \\ \hline
\textbf{OneRec-SAGE} & \underline{0.0651} & \underline{0.0932} & \underline{0.0463} & \underline{0.0557} & \underline{0.0371} & \underline{0.0553} & \underline{0.0256} & \underline{0.0320} & \underline{0.0701} & \underline{0.0960} & \underline{0.0502}  & \underline{0.0585} \\
\textbf{Only SFT (TextRec)} & 0.0291 & 0.0495 & 0.0158 & 0.0204 & 0.0163 & 0.0263 & 0.0104 & 0.0139 & 0.0386 & 0.0571 & 0.0169 & 0.0212 \\
\textbf{TextRec} & \textbf{0.0683} & \textbf{0.0973} & \textbf{0.0497} & \textbf{0.0590} & \textbf{0.0385} & \textbf{0.0575} & \textbf{0.0271} & \textbf{0.0339} & \textbf{0.0738} & \textbf{0.1018} & \textbf{0.0533} & \textbf{0.0622} \\ \hline
\textit{Improve} & \textit{+5.23\%} & \textit{+4.84\%} & \textit{+7.8\%} & \textit{+7.27\%} & \textit{+4.62\%} & \textit{+4.35\%} & \textit{+6.27\%} & \textit{+6.94\%} & \textit{+5.73\%} & \textit{+6.26\%} & \textit{+6.81\%} & \textit{+7.05\%} \\ \hline
\end{tabular}%
}
\end{table*}

\begin{table*}[t]
\centering
\caption{Relative to OneRec‑GBPO, SAGE delivers significant gains in cold‑start recovery and recommendation diversity.}
\label{tab:main_results2}
\resizebox{\textwidth}{!}{%
\begin{tabular}{c|cc|cc|cc}
\hline
\multicolumn{1}{c|}{\textbf{Dataset}} & \multicolumn{2}{c|}{\textbf{Beauty}} & \multicolumn{2}{c|}{\textbf{Sports}} & \multicolumn{2}{c}{\textbf{Toys}} \\ \hline
\multicolumn{1}{c|}{\textbf{Method}} & \textbf{Entropy@10} & \textbf{Cold-Recall@10} & \textbf{Entropy@10} & \textbf{Cold-Recall@10} & \textbf{Entropy@10} & \textbf{Cold-Recall@10} \\ \hline
OneRec-GBPO & 2.285 & 0.326 & 2.274 & 0.282 & 2.315 & 0.365 \\ \hline
\textbf{OneRec-SAGE} & \textbf{2.551} & \textbf{0.619} & \textbf{2.528} & \textbf{0.567} & \textbf{2.574} & \textbf{0.705} \\
\textbf{TextRec} & \underline{2.420} & \underline{0.487} & \underline{2.510} & \underline{0.460} & \underline{2.485} & \underline{0.522} \\ \hline
\textit{Improve} & \textit{+11.64\%} & \textit{+89.9\%} & \textit{+11.17\%} & \textit{+101\%} & \textit{+11.19\%} & \textit{+93.2\%} \\ \hline
\end{tabular}%
}
\end{table*}

\begin{table*}[t]
\centering
\caption{Performance on RecIF-Bench. Bold indicates the best and underline indicates the second best.}
\label{tab:recif_main}
\resizebox{\textwidth}{!}{%
\begin{tabular}{l|ccc|ccc|ccc|c|ccc|cc}
\hline
\multirow{2}{*}{\textbf{Method}} &
\multicolumn{3}{c|}{\textbf{Short Video Rec}} &
\multicolumn{3}{c|}{\textbf{Ad Rec}} &
\multicolumn{3}{c|}{\textbf{Product Rec}} &
\multicolumn{1}{c|}{\textbf{Label Pred.}} &
\multicolumn{3}{c|}{\textbf{Interactive Rec}} &
\multicolumn{2}{c}{\textbf{Label-Cond. Rec}} \\
& Pass@1 & Pass@32 & Recall@32 & Pass@1 & Pass@32 & Recall@32 & Pass@1 & Pass@32 & Recall@32 & AUC & Pass@1 & Pass@32 & Recall@32 & Pass@32 & Recall@32 \\
\hline
SASRec & 0.0045 & 0.1003 & 0.0119 & 0.0044 & 0.0980 & 0.0293 & 0.0052 & 0.0914 & 0.0175 & 0.6244 &  -  &  -  &  -  & 0.0380 & 0.0140 \\
ReaRec & 0.0052 & 0.1002 & 0.0120 & 0.0035 & 0.1054 & 0.0327 & 0.0030 & 0.0907 & 0.0189 & 0.6204 &  -  &  -  &  -  & 0.0381 & 0.0137 \\
TIGER & 0.0168 & 0.1061 & 0.0132 & 0.0125 & 0.1769 & 0.0581 & 0.0120 & 0.1276 & 0.0283 & 0.6675 &  -  &  -  &  -  & 0.0337 & 0.0123 \\
LC-Rec & 0.0341 & 0.1306 & 0.0180 & 0.0197 & 0.2096 & 0.0723 & 0.0178 & 0.1809 & 0.0416 & 0.6139 & 0.0890 & 0.3730 & 0.2394 & 0.0420 & 0.0170 \\
OneRec-GRPO & 0.0548 & 0.2122 & 0.0369 & 0.0259 & 0.2700 & 0.0964 & 0.0223 & 0.2290 & 0.0538 & 0.6912 &  0.1250 & 0.5080 & 0.3458 & 0.0549 & 0.0235 \\
OneRec-GBPO & 0.0551 & 0.2134 & 0.0377 & 0.0264 & 0.2731 & 0.0979 & 0.0227 & 0.2298 & 0.0545 & 0.6918 &  0.1256 & 0.5096 & 0.3481 & 0.0553 & 0.0248 \\
\hline
\textbf{OneRec-SAGE} & \underline{0.0554} & \underline{0.2142} & \underline{0.0381} & \underline{0.0266} & \underline{0.2733} & \underline{0.0983} & \underline{0.0229} & \underline{0.2302} & \underline{0.0554} & \underline{0.6919} & \underline{0.1258} & \underline{0.5107} & \underline{0.3495} & \underline{0.0561} & \underline{0.0254} \\
\textbf{Only SFT (TextRec)} & 0.0049 & 0.0920 & 0.0107 & 0.0041 & 0.0940 & 0.0235 & 0.0032 & 0.0959 & 0.0161 & 0.6028 & 0.0410 & 0.2680 & 0.1660 & 0.0334 & 0.0128 \\
\textbf{TextRec} & \textbf{0.0574} & \textbf{0.2196} & \textbf{0.0392} & \textbf{0.0273} & \textbf{0.2798} & \textbf{0.1015} & \textbf{0.0231} & \textbf{0.2410} & \textbf{0.0568} & \textbf{0.7017} & \textbf{0.1310} & \textbf{0.5280} & \textbf{0.3648} & \textbf{0.0574} & \textbf{0.0261} \\
\hline
\end{tabular}%
}
\end{table*}

\section{Experiments}
\label{sec:Experiments}

We conduct experiments on both public academic benchmarks and a large-scale industrial benchmark to evaluate \textbf{SAGE} as a general-purpose optimizer for list-wise generative recommendation.
Our goal is to verify that SAGE (i) improves accuracy while unlocking cold-start learning signals, (ii) mitigates diversity collapse under rejection-dominated feedback, and (iii) remains numerically stable.
We evaluate SAGE across two representative action spaces: Semantic-ID generation (OneRec-style) and native-text generation (TextRec-style).

\subsection{Experimental Setup}

\subsubsection{Datasets}
\textbf{Amazon Product Reviews.}
We evaluate sequential recommendation on three real-world subsets from the Amazon Product Reviews dataset ~\cite{mcauley2015image}, which contains user reviews and item metadata from May 1996 to July 2014. We use the three categories \textit{Sports}, \textit{Beauty}, and \textit{Toys}.

\begin{table}[t] 
\centering  
\caption{Diversity and cold-start evaluation on RecIF-Bench.}
\label{tab:recif_entropy_cold}
\resizebox{0.48\textwidth}{!}{%
\begin{tabular}{l|cc}
\hline
\multirow{2}{*}{\textbf{Method}} &
\multicolumn{2}{c}{\textbf{Short Video}} \\
& Entropy@10 & Cold-Recall@10 \\
\hline
OneRec-GBPO & 4.284 & 0.521\\
\hline
\textbf{OneRec-SAGE} & \textbf{4.455} & \textbf{0.637} \\
\textbf{TextRec} & \underline{4.317} & \underline{0.594} \\
\hline
\end{tabular}}
\end{table}

\textbf{RecIF-Bench.}
In addition, we evaluate on \textbf{RecIF-Bench} ~\cite{zhou2025openonerec}, an instruction-following benchmark for generative recommendation foundation models. RecIF-Bench contains approximately \textbf{120M} interactions from \textbf{200K} users across three industrial domains: \textit{Short Video} (impression sequences with multi-behavior feedback such as like/follow/comment/effective view/dislike), \textit{Ad} (click sequences), and \textit{Product} (click sequences). The benchmark provides (i) \textbf{user portraits} written as narrative text interleaved with behavioral evidence, (ii) \textbf{item-side metadata and multimodal representations} (text and visual embeddings; dense captions for a large portion of videos), and (iii) \textbf{interaction-side labels} enabling behavior-conditional modeling.

We follow the official \textbf{strict user-based split} protocol: 20\% users are held out for testing with zero user leakage; for each user, the interaction sequence is temporally split into history context and future targets. We report results on the core \textbf{five recommendation tasks} (Short Video Rec, Ad Rec, Product Rec, Interactive Rec, and Label-Conditional Rec) and the pointwise \textbf{Label Prediction} task.

\paragraph{Native-text action space (selection within candidate pool).}
In TextRec, the policy does not freely generate arbitrary item strings. Instead, for each context $c$ we provide a candidate pool $\mathcal{I}(c)$ with textual item renderings (e.g., title/tags/category keywords), and the model outputs a length-$L$ slate by selecting items from $\mathcal{I}(c)$ (with deterministic parsing and validity checks). Outputs that do not match any candidate are treated as invalid and discarded (or mapped to a special null action), ensuring a well-defined and reproducible action space.

\subsubsection{Baselines} To validate the effectiveness of our approach, we compare against three groups of competitive baselines, covering classical discriminative sequential recommenders, recent generative/retrieval-augmented recommenders, and OneRec-family variants.

\textbf{Discriminative sequential recommenders:} SASRec~\cite{kang2018self}. These methods are inherently task-specific, and thus require training a separate model for each RecIF task.

\textbf{Generative / retrieval-augmented recommenders:} ReaRec~\cite{tang2025think}, TIGER~\cite{rajput2023recommender}, and LC-Rec~\cite{zheng2024adapting}. Following prior practice, we adapt these methods with task-specific modifications to support different RecIF tasks. For a fair comparison at the foundation-model scale, we implement LC-Rec as \textbf{LC-Rec-8B} by instantiating it with a comparable \textbf{Qwen3-8B} backbone.

\textbf{OneRec variants and our method:} OneRec-GRPO~\cite{zhou2025openonerec} (Semantic IDs + GRPO), OneRec-GBPO~\cite{zhou2025onerec} (Semantic IDs + GBPO), OneRec-SAGE (applying SAGE on the Semantic-ID architecture), \textbf{Only SFT (TextRec)} (native-vocabulary model trained only with supervised fine-tuning), and \textbf{TextRec} (native-vocabulary model further optimized with SAGE).

\subsubsection{Evaluation Metrics}
\textbf{Amazon metrics.}
We evaluate top-$K$ performance (default $K=10$) from three perspectives:
(i) \textbf{Accuracy} using NDCG@K and Recall@K;
(ii) \textbf{Diversity} using Entropy@K. Since all items in a dataset category share the same root label (e.g., ''Beauty''), entropy at the root level is trivial; therefore, we compute Shannon Entropy using second-level sub-categories from item metadata (e.g., \textit{Skin Care}, \textit{Makeup}, \textit{Fragrance});
(iii) \textbf{Cold-start} using Cold-Recall@K, where \textit{cold items} are defined as the bottom 5\% of the interaction-frequency distribution in the training set.

\textbf{RecIF-Bench metrics.}
We follow the official RecIF-Bench evaluation ~\cite{zhou2025openonerec}. For recommendation tasks, we using \textbf{Pass@K} and \textbf{Recall@K} with $K\in\{1,32\}$. For \textit{Label Prediction}, we using \textbf{AUC}.

\subsubsection{Implementation Details}
\textbf{Backbone and generation.}
The base model is Qwen3-8B. We set slate length to $L=10$.

\textbf{Training recipe.}
We use a two-stage pipeline (SFT then RLHF) as described in Section~\ref{sec:Method}. The SFT stage is trained for 1 epoch. In the RLHF stage, we use AdamW with a learning rate of $5e-6$. For SAGE hyperparameters, based on validation tuning, we set the Positive Boost Factor $\epsilon_{boost}=0.3$ and Diversity Temperature $\beta=0.5$.

\textbf{RecIF-Bench specific settings.}
For TextRec, items are represented purely by \textbf{native-text} (title/tags/category keywords) without introducing a separate item-token vocabulary. For interactive and conditional recommendation, $L=10$, and we generate candidates using nucleus sampling with temperature 0.9. For OneRec-GRPO, OneRec-GBPO and OneRec-SAGE, we use the public checkpoints and evaluation settings in OpenOneRec ~\cite{zhou2025openonerec}. For task-specific sequential baselines and TIGER, we reuse the reported benchmark numbers under the same protocol.

\subsection{Main Results}

\subsubsection{Results on Amazon Product Reviews}
Table~\ref{tab:main_results1} shows that \textbf{TextRec consistently achieves the best top-$K$ accuracy} across Beauty, Sports, and Toys. Compared to the strongest baseline \textbf{OneRec-GBPO}, TextRec improves NDCG by \textbf{6.27\% to 7.8\%}, indicating that directly reusing the native LLM vocabulary—when paired with preference optimization—can outperform architectures relying on separate Semantic-ID tokenizers.

The \textbf{Only SFT (TextRec)} variant performs poorly and lags behind SASRec, suggesting that SFT mainly teaches formatting and coarse preference patterns; without preference optimization, the model fails to align generation with ranking objectives. Applying SAGE to the Semantic-ID architecture (\textbf{OneRec-SAGE}) yields competitive but mixed accuracy changes, implying that aggressive exploration is harder to stabilize when the output space is constrained by a discrete Semantic-ID vocabulary.

Beyond accuracy, Table~\ref{tab:main_results2} highlights SAGE’s effect on long-tail discovery and diversity. Relative to OneRec-GBPO, SAGE yields large gains in \textbf{Cold-Recall} and also improves \textbf{Entropy}, supporting our diagnosis that GBPO-style symmetric clipping tends to suppress positive learning signals for rare items and encourages conservative, popularity-biased updates.

\subsubsection{Results on RecIF-Bench}
Table~\ref{tab:recif_main} reports results on six RecIF-Bench tasks. \textbf{TextRec consistently outperforms OneRec-GBPO} across the core recommendation tasks, with improvements that are especially clear on instruction-conditioned settings (e.g., \textbf{Interactive Rec} and \textbf{Label-Conditional Rec}). This indicates that representing items with \textbf{native-text} can better exploit the backbone LLM’s semantic priors when the task is framed as instruction following rather than pure next-item prediction.

The \textbf{Only SFT} variant again underperforms, consistent with the Amazon findings: SFT alone does not provide sufficiently strong preference signals to drive high-quality ranking under the RecIF-Bench protocol. \textbf{OneRec-SAGE} remains competitive but does not consistently dominate top-$K$ accuracy, reinforcing the empirical pattern that exploration can become over-aggressive when the action space is limited by a Semantic-ID vocabulary.

Finally, Table~\ref{tab:recif_entropy_cold} evaluates \textbf{diversity and cold-start} on RecIF-Bench (Short Video). Relative to OneRec-GBPO, TextRec improves both Entropy@10 and Cold-Recall@10, while OneRec-SAGE achieves the highest absolute values on these two metrics. Taken together with the accuracy table, this suggests a consistent trade-off: SAGE increases exploration and long-tail coverage; the native-vocabulary TextRec setting better preserves accuracy under such exploration than Semantic-ID generation.

\begin{figure}[t]
    \centering
    \includegraphics[width=0.9\linewidth]{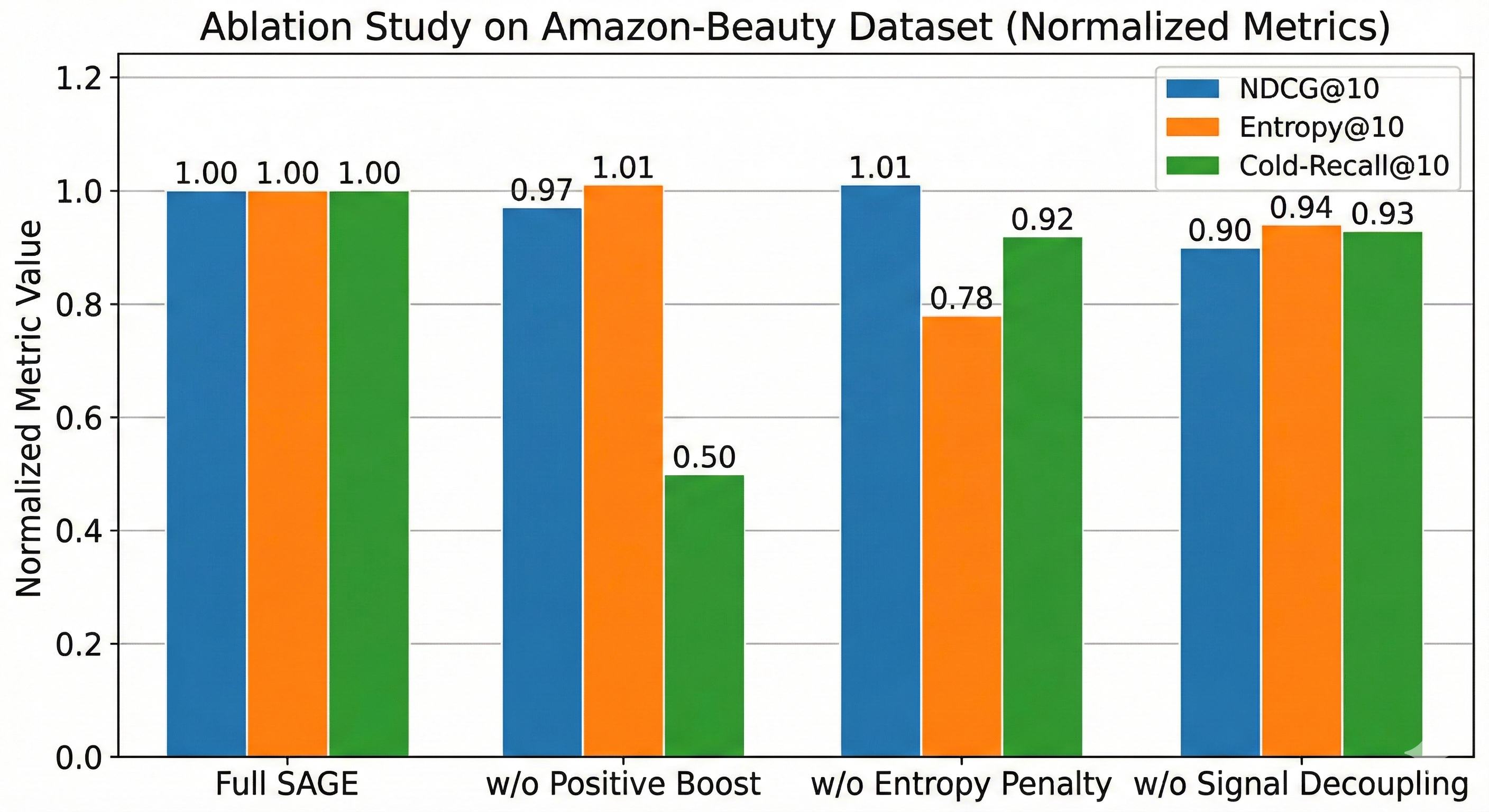} 
    \caption{\textbf{Ablation Study on Amazon-Beauty Dataset.} Metrics are normalized to the full SAGE model. Removing individual components leads to degradation in their respective targeted areas (e.g., removing Boost hurts Cold-Start most), validating the design of each module.}
    \label{fig:ablation}
\end{figure}

\subsection{Ablation Study}
To isolate the contribution of each component, we conduct ablations on \textbf{Amazon-Beauty}. Figure~\ref{fig:ablation} normalizes metrics to the full model.

Removing \textbf{Positive Boost} ($\epsilon_{boost}=0$) drastically reduces Cold-Start Recall, showing that relaxing the conservative positive update boundary is crucial for long-tail discovery. Removing the \textbf{Entropy Penalty} substantially decreases Entropy@10, indicating that diversity-aware negative feedback prevents collapse into redundant popular outputs. Replacing \textbf{Signal Decoupling} with a simple weighted reward sum reduces NDCG@10 and other metrics, supporting the hypothesis that multi-objective reward mixing can cause signal interference and unstable optimization. Overall, the best results require all three components working together.

\section{Conclusion}
\label{sec:Conclusion}

We studied optimization failures in generative recommendation, focusing on the "Symmetric Conservatism'' induced by GBPO-style bounded updates, which empirically manifests as cold-start suppression and diversity collapse. To address this in list-wise generation, we proposed \textbf{SAGE (Sequence-level Adaptive Gradient Evolution)}, featuring (i) \textbf{Asymmetric Adaptive Dynamics} via a Positive Boost for high-potential successful slates and an Entropy-Aware Penalty for low-diversity failures, and (ii) \textbf{Multi-Objective Signal Decoupling} to preserve gradient fidelity under mixed feedback.

Across three Amazon Product Reviews datasets and the large-scale industrial \textbf{RecIF-Bench} benchmark, \textbf{TextRec}—which reuses the LLM’s \textbf{native vocabulary} and is optimized with SAGE—consistently achieves the strongest overall accuracy while substantially improving long-tail coverage and recommendation diversity relative to OneRec-GBPO. The results also reveal a practical stability pattern: applying strong exploration to a constrained Semantic-ID action space can increase diversity but may sacrifice top-$K$ accuracy, whereas native-text generation better maintains the accuracy--diversity balance under SAGE.

A remaining limitation is that native-vocabulary recommendation requires language-level item rendering, which can be challenging for latency-sensitive systems; nevertheless, the consistent gains on both academic and industrial benchmarks suggest that native-text generation is a promising direction as compute and deployment tooling improve.

\clearpage

\bibliographystyle{ACM-Reference-Format}
\bibliography{ref.bib}

@STRING{IJCAI   = {Proc. Int. Joint Conf. Artificial Intell.}}

@article{li2025survey,
  title={A Survey of Generative Recommendation from a Tri-Decoupled Perspective: Tokenization, Architecture, and Optimization},
  author={Li, Xiaopeng and Chen, Bo and She, Junda and Cao, Shiteng and Wang, You and Jia, Qinlin and He, Haiying and Zhou, Zheli and Liu, Zhao and Liu, Ji and others},
  year={2025},
  publisher={Preprints}
}

@article{deng2025onerec,
  title={Onerec: Unifying retrieve and rank with generative recommender and iterative preference alignment},
  author={Deng, Jiaxin and Wang, Shiyao and Cai, Kuo and Ren, Lejian and Hu, Qigen and Ding, Weifeng and Luo, Qiang and Zhou, Guorui},
  journal={arXiv preprint arXiv:2502.18965},
  year={2025}
}

@article{zhou2025onerec,
  title={Onerec-v2 technical report},
  author={Zhou, Guorui and Hu, Hengrui and Cheng, Hongtao and Wang, Huanjie and Deng, Jiaxin and Zhang, Jinghao and Cai, Kuo and Ren, Lejian and Ren, Lu and Yu, Liao and others},
  journal={arXiv preprint arXiv:2508.20900},
  year={2025}
}

@article{zhou2025openonerec,
  title={OpenOneRec Technical Report},
  author={Zhou, Guorui and Bao, Honghui and Huang, Jiaming and Deng, Jiaxin and Zhang, Jinghao and She, Junda and Cai, Kuo and Ren, Lejian and Ren, Lu and Luo, Qiang and others},
  journal={arXiv preprint arXiv:2512.24762},
  year={2025}
}

@article{yu2025dapo,
  title={Dapo: An open-source llm reinforcement learning system at scale},
  author={Yu, Qiying and Zhang, Zheng and Zhu, Ruofei and Yuan, Yufeng and Zuo, Xiaochen and Yue, Yu and Dai, Weinan and Fan, Tiantian and Liu, Gaohong and Liu, Lingjun and others},
  journal={arXiv preprint arXiv:2503.14476},
  year={2025}
}

@article{xi2025bapo,
  title={Bapo: Stabilizing off-policy reinforcement learning for llms via balanced policy optimization with adaptive clipping},
  author={Xi, Zhiheng and Guo, Xin and Nan, Yang and Zhou, Enyu and Shen, Junrui and Chen, Wenxiang and Liu, Jiaqi and Huang, Jixuan and Zhang, Zhihao and Guo, Honglin and others},
  journal={arXiv preprint arXiv:2510.18927},
  year={2025}
}

@article{zheng2025group,
  title={Group sequence policy optimization},
  author={Zheng, Chujie and Liu, Shixuan and Li, Mingze and Chen, Xiong-Hui and Yu, Bowen and Gao, Chang and Dang, Kai and Liu, Yuqiong and Men, Rui and Yang, An and others},
  journal={arXiv preprint arXiv:2507.18071},
  year={2025}
}

@article{liu2026gdpo,
  title={GDPO: Group reward-Decoupled Normalization Policy Optimization for Multi-reward RL Optimization},
  author={Liu, Shih-Yang and Dong, Xin and Lu, Ximing and Diao, Shizhe and Belcak, Peter and Liu, Mingjie and Chen, Min-Hung and Yin, Hongxu and Wang, Yu-Chiang Frank and Cheng, Kwang-Ting and others},
  journal={arXiv preprint arXiv:2601.05242},
  year={2026}
}

@article{koren2009matrix,
  title={Matrix factorization techniques for recommender systems},
  author={Koren, Yehuda and Bell, Robert and Volinsky, Chris},
  journal={Computer},
  volume={42},
  number={8},
  pages={30--37},
  year={2009},
  publisher={IEEE}
}

@inproceedings{covington2016deep,
  title={Deep neural networks for youtube recommendations},
  author={Covington, Paul and Adams, Jay and Sargin, Emre},
  booktitle={Proceedings of the 10th ACM conference on recommender systems},
  pages={191--198},
  year={2016}
}

@inproceedings{mcauley2015image,
  title={Image-based recommendations on styles and substitutes},
  author={McAuley, Julian and Targett, Christopher and Shi, Qinfeng and Van Den Hengel, Anton},
  booktitle={Proceedings of the 38th international ACM SIGIR conference on research and development in information retrieval},
  pages={43--52},
  year={2015}
}

@article{wei2022chain,
  title={Chain-of-thought prompting elicits reasoning in large language models},
  author={Wei, Jason and Wang, Xuezhi and Schuurmans, Dale and Bosma, Maarten and Xia, Fei and Chi, Ed and Le, Quoc V and Zhou, Denny and others},
  journal={Advances in neural information processing systems},
  volume={35},
  pages={24824--24837},
  year={2022}
}

@article{afsar2022reinforcement,
  title={Reinforcement learning based recommender systems: A survey},
  author={Afsar, M Mehdi and Crump, Trafford and Far, Behrouz},
  journal={ACM Computing Surveys},
  volume={55},
  number={7},
  pages={1--38},
  year={2022},
  publisher={ACM New York, NY}
}

@inproceedings{ie2019slateq,
  title={SlateQ: A Tractable Decomposition for Reinforcement Learning with Recommendation Sets.},
  author={Ie, Eugene and Jain, Vihan and Wang, Jing and Narvekar, Sanmit and Agarwal, Ritesh and Wu, Rui and Cheng, Heng-Tze and Chandra, Tushar and Boutilier, Craig},
  booktitle={IJCAI},
  volume={19},
  pages={2592--2599},
  year={2019}
}

@inproceedings{chen2019top,
  title={Top-k off-policy correction for a REINFORCE recommender system},
  author={Chen, Minmin and Beutel, Alex and Covington, Paul and Jain, Sagar and Belletti, Francois and Chi, Ed H},
  booktitle={Proceedings of the twelfth ACM international conference on web search and data mining},
  pages={456--464},
  year={2019}
}

@article{zhao2017deep,
  title={Deep reinforcement learning for list-wise recommendations},
  author={Zhao, Xiangyu and Zhang, Liang and Xia, Long and Ding, Zhuoye and Yin, Dawei and Tang, Jiliang},
  journal={arXiv preprint arXiv:1801.00209},
  year={2017}
}

@inproceedings{artetxe2020cross,
  title={On the cross-lingual transferability of monolingual representations},
  author={Artetxe, Mikel and Ruder, Sebastian and Yogatama, Dani},
  booktitle={Proceedings of the 58th annual meeting of the association for computational linguistics},
  pages={4623--4637},
  year={2020}
}

@article{ranzato2015sequence,
  title={Sequence level training with recurrent neural networks},
  author={Ranzato, Marc'Aurelio and Chopra, Sumit and Auli, Michael and Zaremba, Wojciech},
  journal={arXiv preprint arXiv:1511.06732},
  year={2015}
}

@article{kojima2022large,
  title={Large language models are zero-shot reasoners},
  author={Kojima, Takeshi and Gu, Shixiang Shane and Reid, Machel and Matsuo, Yutaka and Iwasawa, Yusuke},
  journal={Advances in neural information processing systems},
  volume={35},
  pages={22199--22213},
  year={2022}
}

@article{brown2020language,
  title={Language models are few-shot learners},
  author={Brown, Tom and Mann, Benjamin and Ryder, Nick and Subbiah, Melanie and Kaplan, Jared D and Dhariwal, Prafulla and Neelakantan, Arvind and Shyam, Pranav and Sastry, Girish and Askell, Amanda and others},
  journal={Advances in neural information processing systems},
  volume={33},
  pages={1877--1901},
  year={2020}
}

@inproceedings{devlin2019bert,
  title={Bert: Pre-training of deep bidirectional transformers for language understanding},
  author={Devlin, Jacob and Chang, Ming-Wei and Lee, Kenton and Toutanova, Kristina},
  booktitle={Proceedings of the 2019 conference of the North American chapter of the association for computational linguistics: human language technologies, volume 1 (long and short papers)},
  pages={4171--4186},
  year={2019}
}

@article{rajput2023recommender,
  title={Recommender systems with generative retrieval},
  author={Rajput, Shashank and Mehta, Nikhil and Singh, Anima and Hulikal Keshavan, Raghunandan and Vu, Trung and Heldt, Lukasz and Hong, Lichan and Tay, Yi and Tran, Vinh and Samost, Jonah and others},
  journal={Advances in Neural Information Processing Systems},
  volume={36},
  pages={10299--10315},
  year={2023}
}

@inproceedings{zheng2024adapting,
  title={Adapting large language models by integrating collaborative semantics for recommendation},
  author={Zheng, Bowen and Hou, Yupeng and Lu, Hongyu and Chen, Yu and Zhao, Wayne Xin and Chen, Ming and Wen, Ji-Rong},
  booktitle={2024 IEEE 40th International Conference on Data Engineering (ICDE)},
  pages={1435--1448},
  year={2024},
  organization={IEEE}
}

@article{shao2024deepseekmath,
  title={Deepseekmath: Pushing the limits of mathematical reasoning in open language models},
  author={Shao, Zhihong and Wang, Peiyi and Zhu, Qihao and Xu, Runxin and Song, Junxiao and Bi, Xiao and Zhang, Haowei and Zhang, Mingchuan and Li, YK and Wu, Y and others},
  journal={arXiv preprint arXiv:2402.03300},
  year={2024}
}

@inproceedings{kang2018self,
  title={Self-attentive sequential recommendation},
  author={Kang, Wang-Cheng and McAuley, Julian},
  booktitle={2018 IEEE international conference on data mining (ICDM)},
  pages={197--206},
  year={2018},
  organization={IEEE}
}

@article{tang2025think,
  title={Think before recommend: Unleashing the latent reasoning power for sequential recommendation},
  author={Tang, Jiakai and Dai, Sunhao and Shi, Teng and Xu, Jun and Chen, Xu and Chen, Wen and Wu, Jian and Jiang, Yuning},
  journal={arXiv preprint arXiv:2503.22675},
  year={2025}
}

@inproceedings{liu2017cascade,
  title={Cascade ranking for operational e-commerce search},
  author={Liu, Shichen and Xiao, Fei and Ou, Wenwu and Si, Luo},
  booktitle={Proceedings of the 23rd ACM SIGKDD International Conference on Knowledge Discovery and Data Mining},
  pages={1557--1565},
  year={2017}
}

@inproceedings{qin2022rankflow,
  title={Rankflow: Joint optimization of multi-stage cascade ranking systems as flows},
  author={Qin, Jiarui and Zhu, Jiachen and Chen, Bo and Liu, Zhirong and Liu, Weiwen and Tang, Ruiming and Zhang, Rui and Yu, Yong and Zhang, Weinan},
  booktitle={Proceedings of the 45th International ACM SIGIR Conference on Research and Development in Information Retrieval},
  pages={814--824},
  year={2022}
}

@article{luo2024integrating,
  title={Integrating large language models into recommendation via mutual augmentation and adaptive aggregation},
  author={Luo, Sichun and Yao, Yuxuan and He, Bowei and Huang, Yinya and Zhou, Aojun and Zhang, Xinyi and Xiao, Yuanzhang and Zhan, Mingjie and Song, Linqi},
  journal={arXiv preprint arXiv:2401.13870},
  year={2024}
}

@article{li2023gpt4rec,
  title={GPT4Rec: A generative framework for personalized recommendation and user interests interpretation},
  author={Li, Jinming and Zhang, Wentao and Wang, Tian and Xiong, Guanglei and Lu, Alan and Medioni, Gerard},
  journal={arXiv preprint arXiv:2304.03879},
  year={2023}
}

@article{yang2023palr,
  title={Palr: Personalization aware llms for recommendation},
  author={Yang, Fan and Chen, Zheng and Jiang, Ziyan and Cho, Eunah and Huang, Xiaojiang and Lu, Yanbin},
  journal={arXiv preprint arXiv:2305.07622},
  year={2023}
}

@inproceedings{liao2024llara,
  title={Llara: Large language-recommendation assistant},
  author={Liao, Jiayi and Li, Sihang and Yang, Zhengyi and Wu, Jiancan and Yuan, Yancheng and Wang, Xiang and He, Xiangnan},
  booktitle={Proceedings of the 47th International ACM SIGIR Conference on Research and Development in Information Retrieval},
  pages={1785--1795},
  year={2024}
}

@inproceedings{ji2024genrec,
  title={Genrec: Large language model for generative recommendation},
  author={Ji, Jianchao and Li, Zelong and Xu, Shuyuan and Hua, Wenyue and Ge, Yingqiang and Tan, Juntao and Zhang, Yongfeng},
  booktitle={European Conference on Information Retrieval},
  pages={494--502},
  year={2024},
  organization={Springer}
}


\end{document}